\pdfoutput=1

\documentclass[11pt]{article}

\usepackage[]{acl}

\usepackage{times}
\usepackage{latexsym}

\usepackage[T1]{fontenc}

\usepackage[utf8]{inputenc}

\usepackage{microtype}

\usepackage{inconsolata}
\usepackage{algorithm}
\usepackage{algpseudocode}
\usepackage{enumitem}
\usepackage{float}
\usepackage{amsmath}
\usepackage{graphicx}
\usepackage{booktabs}
\usepackage{xspace}
\usepackage{multirow}
\usepackage{tabularx}
\usepackage{hyperref}
\usepackage{geometry} 
\usepackage[bottom, hang, flushmargin]{footmisc} 
\newcommand{\dataset}{ItineraryBench\xspace}

\title{TDAG: A Multi-Agent Framework based on Dynamic Task Decomposition and Agent Generation}

\author{
Yaoxiang Wang$^{\clubsuit\diamondsuit}$\thanks{Work done during an internship at Shanghai AI Laboratory}, 
Zhiyong Wu$^{\diamondsuit}$\thanks{Correspondence to: Jinsong Su, Zhiyong Wu.}, 
Junfeng Yao$^{\clubsuit}$, 
Jinsong Su$^{\clubsuit\dagger}$
\\
$^\clubsuit$Xiamen University \quad
$^\diamondsuit$Shanghai AI Laboratory\\
\texttt{jssu@xmu.edu.cn, wuzhiyong@pjlab.org.cn}
}

\begin{document}
\maketitle

\footnotetext{The model, data and code are available at \url{https://github.com/yxwang8775/TDAG}.}

\begin{abstract}
The emergence of Large Language Models (LLMs) like ChatGPT has inspired the development of LLM-based agents capable of addressing complex, real-world tasks. However, these agents often struggle during task execution due to methodological constraints, such as error propagation and limited adaptability. To address this issue, we propose a multi-agent framework based on dynamic Task Decomposition and Agent Generation (TDAG). This framework dynamically decomposes complex tasks into smaller subtasks and assigns each to a specifically generated subagent, thereby enhancing adaptability in diverse and unpredictable real-world tasks. 
Simultaneously, existing benchmarks often lack the granularity needed to evaluate incremental progress in complex, multi-step tasks. In response, we introduce \dataset in the context of travel planning, featuring interconnected, progressively complex tasks with a fine-grained evaluation system. \dataset is designed to assess agents' abilities in memory, planning, and tool usage across tasks of varying complexity. Our experimental results reveal that TDAG significantly outperforms established baselines, showcasing its superior adaptability and context awareness in complex task scenarios.
\end{abstract}

\section{Introduction}
The emergence of Large Language Models (LLMs)~\cite{opt,gpt4}, such as ChatGPT, represents a noteworthy milestone in artificial intelligence, laying the groundwork for the creation of LLM-based agents~\cite{autogpt, babyagi} with the capacity to automate tasks on behalf of humans.

Despite advancements, LLM-based agents face substantial challenges in real-world tasks that demand complex planning, multi-stage reasoning, and tool utilization~\cite{agent_reasoning, tool_learning}. For instance, in a recent agent benchmark~\cite{gaia}, even GPT4 falls short with a success rate of 14\%, while humans effortlessly exceed 92\%. Empowering agents to effectively address real-world tasks is challenging, arising from difficulties in both benchmark and methodological perspectives.

From a benchmark perspective, we have witnessed lots of agent-related benchmarks in the passing year, covering domains like web manipulation~\cite{webshop}, gaming~\cite{voyager}, and software control~\cite{sheetcopilot}. However, the evaluation metrics in current benchmarks often lack the granularity needed to accurately reflect the incremental progress of agents. In complex multi-step tasks, even if an agent fails to achieve the ultimate targets, it might successfully accomplish some of the subtasks. It is essential to include these partial successes in evaluation metrics to faithfully capture the capabilities of agents. Reporting only a binary score (success or failure) can lead to misconceptions such as emergent abilities~\cite{emergent}.

From a methodological perspective, existing approaches~\cite{planandexecute,adapt,planandsolve} aim to break down a complex request into a sequence of subtasks to reduce complexity and address the challenge of excessively long input. However, they encounter two notable drawbacks: error propagation and limited adaptability. In many studies, after the decomposition of the problem, all subtasks become fixed and unalterable. In such a setup, if an early subtask encounters failure, the error propagates, resulting in the failure of the entire task. Regarding subtask execution, prevailing efforts often involve the manual construction of hard-coded subagents. This static design tends to lack generality and is unable to scale effectively to handle diverse real-world tasks, given its labor and time-intensive nature. 

To bridge the gap and facilitate research on complex reasoning and planning in LLM-based agents, we have made two contributions. First, we developed a benchmark that supports fine-grained evaluation of agents' progress. Concurrently, we developed a multi-agent framework for real-world problem-solving. This framework distinguishes itself through dynamic task decomposition and automatic subagent creation.

Our benchmark focuses on agent-assisted travel planning. Concretely, agents in our benchmark are required to act as personal assistants, leveraging computer tools like the databases and the code interpreter to execute tasks. To provide a holistic evaluation, our benchmark encompasses a spectrum of tasks, each
varying in complexity. We also develop a simulator to mimic dynamic real-world scenarios, encompassing the entire pipeline of travel planning—from ticket booking to route/time planning. With the simulator, we are able to assess agents' partial task completion and deliver a nuanced score, which offers a more accurate and fair reflection of agents’ capabilities.

Our multi-agent framework decomposes complex tasks into smaller, more manageable subtasks, while dynamically adjusting subsequent subtasks based on the completion status of preceding ones. For each subtask, a subtask-tailored subagent is generated using LLMs, which differs our design from conventional approaches that rely on a single agent or a set of manually designed subagents. Furthermore, these subagents are equipped with an evolving and adaptable skill library, designed to meet the challenges of varied and unpredictable environments.

Our contributions are summarized as follows:
\begin{itemize}
\item We develop \dataset, featuring interconnected, progressively complex tasks and a fine-grained evaluation system.

\item We introduce a multi-agent framework based on dynamic Task Decomposition and Agent Generation (TDAG), which decomposes a complex task into smaller subtasks, each managed by a specifically generated subagent.

\item We conduct experiments on \dataset, validating the effectiveness of the TDAG framework and highlighting the benefits of fine-grained evaluation.

\end{itemize}
\section{Related Work}

\subsection{Large Language Model-based Agents.}Large language models (LLMs) have demonstrated exceptional capabilities in language understanding, cognition, and reasoning~\cite{opt, instructgpt, gpt4}, inspiring the development of LLM-based agents~\cite{voyager, generative_agents, corex}. These agents integrate LLMs as their core processing units, enabling them to efficiently tackle complex tasks~\cite{cot_agent,agent_survey}. Their widespread application in various domains, from software development to scientific research, demonstrates their versatility~\cite{software_agent,research,oscopilot}. This versatility is further exemplified by innovative open-source projects such as AutoGPT~\cite{autogpt}, BabyAGI~\cite{babyagi}, and Xagent~\cite{xagent}, which leverage LLMs for automated task execution by simply being provided with a task description and a set of available tools.

\subsection{Enhancing Agent Planning with Large Language Models} To enhance the reasoning and planning capabilities of LLM-based agents, various techniques like Chain-of-Thought prompting and task decomposition are employed. Typically, ReAct~\cite{react} uses an iterative process where a LLM generates thoughts and actions based on current observations until task completion. Reflexion~\cite{reflexion} introduces a self-reflection mechanism to improve reasoning quality. Meanwhile, ADAPT~\cite{adapt} employs a recursive strategy that breaks down tasks into subtasks and allows further decomposition when necessary.
We extend these capabilities further by introducing a multi-agent framework where the task is dynamically decomposed and each subtask is then assigned to specially generated subagent.
This approach allows for more flexible and efficient task management, especially in complex and unpredictable scenarios.

\subsection{Evaluating LLM-based Agents.} Recently, various benchmarks have been developed to evaluate the foundational abilities of these agents, such as decision-making, tool utilization, and embodied action capabilities~\cite{decision_dialogue, tool_learning, interaction, gentopia}. Additionally, there has been a focus on evaluating agents in specific real-world scenarios. For instance, AgentBench~\cite{agentbench} has compiled a series of tasks set in multiple real-world scenarios to assess the performance of agents in completing a variety of tasks, including code, web~\cite{webshop}, and game~\cite{alfworld}. Beyond scenario-specific evaluations, there is a growing interest in assessing the comprehensive abilities of agents. OpenAGI~\cite{openagi} is an open-source AGI research and development platform designed for solving multi-step, real-world tasks. GAIA~\cite{gaia} introduces hundreds of real-world questions that necessitate fundamental abilities like reasoning, multi-modality handling. Different from existing studies, \dataset uniquely assesses partial task completions, offering a more refined assessment framework for LLM-based agents. \dataset focuses on travel planning, a scenario that offers a realistic and dynamic environment where tasks vary in complexity and require an agent to demonstrate abilities such as memory, planning, and tool usage, making it an ideal scenario to test agents' adaptability and incremental progress in completing complex, multi-step tasks.

\section{\dataset}
\label{sec:benchmark}
In this section, we first provide an overview of \dataset (Section~\ref{sec:benchmark-overview}), then explain the dataset construction (Section~\ref{sec:dataset-construction}), discuss the multi-level evaluation metrics (Section~\ref{sec:evaluation-metrics}), and highlight key features of our benchmark (Section~\ref{sec:features}).

\subsection{Overview}
\label{sec:benchmark-overview}
We introduce \dataset, a benchmark designed for evaluating agents in complex real-world scenarios, with a particular focus on travel planning. The core task for agents is to plan a trip that involves visits to multiple cities and attractions, while adhering to various constraints such as time and budget. This requires agents to interact with tools to access detailed information about transportation, accommodation, and attractions, as well as to facilitate planning.

\begin{figure}[ht]
\centering
\includegraphics[width=\linewidth]{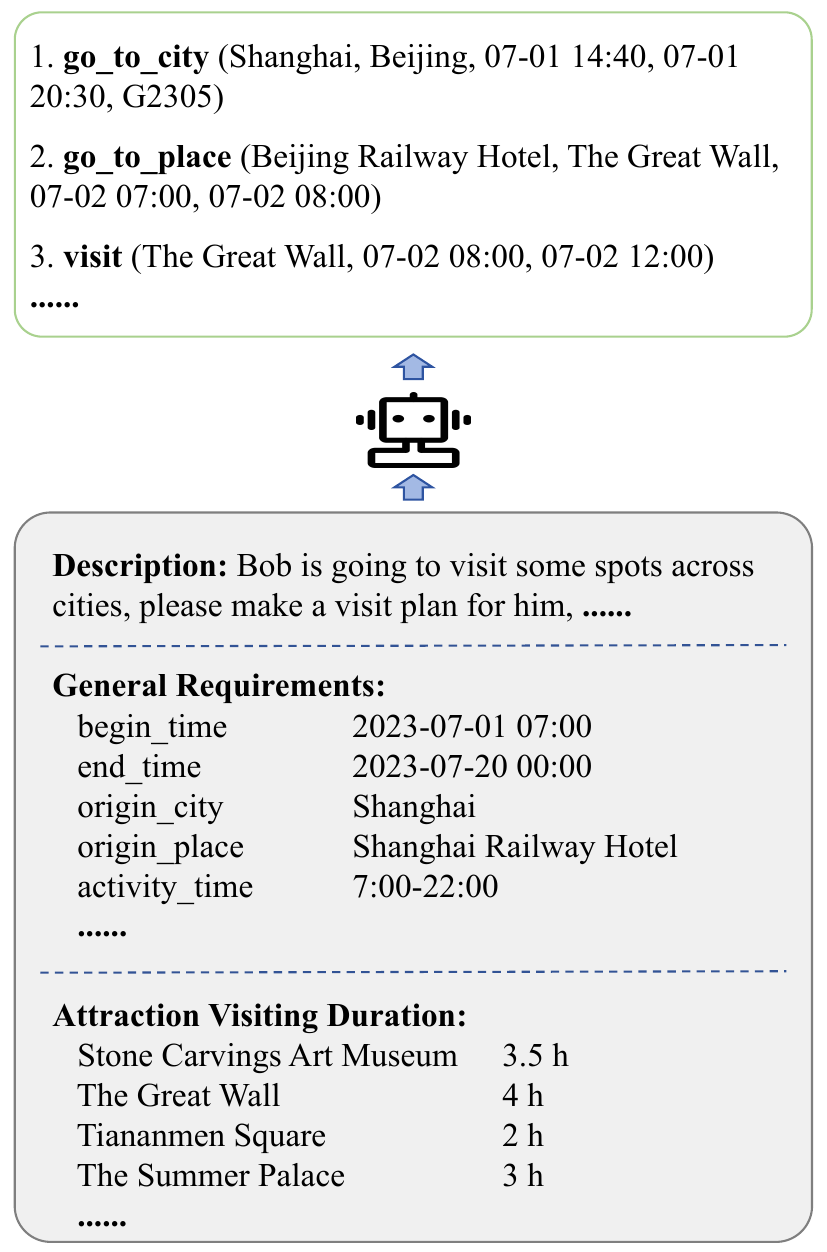}
\caption{\textbf{Example of a travel planning task.} The task specifies initial conditions and constraints for generating a travel itinerary.}
\label{fig:data_example}
\end{figure}

As illustrated in Figure~\ref{fig:data_example}, agents receive a task and are expected to produce a detailed itinerary that outlines activities for each segment of the trip. Formally, the itinerary comprises a sequence of predefined actions, including moving between places and cities, visiting spots, and staying in cities, as summarized in Table~\ref{tab:actions}.

\begin{table*}[h]
\centering
\small
\caption{Actions for itinerary construction.}
\renewcommand{\arraystretch}{1.2} 
\begin{tabular}{|p{0.57\linewidth}|p{0.37\linewidth}|}
\hline
\textbf{Action} & \textbf{Description} \\
\hline
\textbf{go\_to\_place} (origin, destination, depart\_time, arrive\_time) & Travel from one place to another within a city. \\
\textbf{visit} (place, begin\_time, end\_time) & Visit a tourist spot for a specified duration.\\
\textbf{go\_to\_city} (origin, destination, depart\_time, arrive\_time, ticket) & Travel from one city to another. \\
\textbf{stay\_in} (city, begin\_time, end\_time) & Stay in a city for a specified duration. \\
\hline
\end{tabular}

\label{tab:actions}
\end{table*}

\subsection{Dataset Construction}
\label{sec:dataset-construction}

We gathered information of 83 tourist attractions across 15 cities, 846 intra-city and 3903 inter-city transportation routes. Based on this foundation, we employed a combination of manual curation and code-assisted method to construct 364 distinct samples, exclusively reserved as the test set. Each sample represents a unique travel planning scenario, encompassing various combinations of cities, attractions, transportation modes, and scheduling constraints. Detailed statistics can be found in Appendix~\ref{sec:appendix}.

Our benchmark encompasses three distinct task types, each anchored by a central storyline from which numerous data samples emerge. These types provide a comprehensive assessment of the agents' capabilities in handling varying complexities and real-world variables.

\subsubsection*{Type 1: Inter-city Travel Planning}
The primary storyline of this type involves the user planning a journey that spans multiple cities, focusing on determining the order of cities to visit, the duration of stays, and the inter-city transportation. The final itinerary involves \texttt{`go\_to\_city'} and \texttt{`stay\_in'} actions. There are variations in cities to visit, lengths of stay, and available activity hours, etc.

\subsubsection*{Type 2: Intra-city Travel Planning}
In this type of task, the user starts from a fixed location and plans to visit various attractions within a city. The final itinerary involves \texttt{`go\_to\_place'} and \texttt{`visit'} actions, incorporating constraints such as limited attraction opening hours and designated time allocations for meals and activities.

\subsubsection*{Type 3: Combined Intra- and Inter-city Travel Planning}

This type of task is a composite of Types 1 and 2, involving planning visits to attractions within multiple cities. The itinerary involves \texttt{`go\_to\_city'}, \texttt{`go\_to\_place'}, and \texttt{`visit'} actions, representing a more complex version of travel planning.

\noindent\textbf{Tools.} To facilitate these tasks, agents have access to a database and a Python interpreter environment. The database contains information on transportation and attractions across cities and the Python interpreter can be used for calculation and numerical optimization

\subsection{Evaluation Metrics}
\label{sec:evaluation-metrics}
To comprehensively evaluate the performance of agents, we develop a simulated environment that closely resembles real-world scenarios, assessing agents at three levels: Executability, Constraint Satisfaction, and Time and Cost Efficiency. Unlike traditional benchmarks that rely on a singular ground truth, we do not require a predefined ground truth. Instead, agents’ plans are executed within the simulator, which provides real-time feedback by validating the correctness of the actions performed. This allows us to directly score the plans based on their performance within the environment, ensuring that multiple correct answers are accommodated.

\textbf{Level 1 - Executability.} Itineraries are evaluated on their basic feasibility. For example, \texttt{`go\_to\_city(Shanghai, Beijing, 07-01 14:40, 07-01 20:30, G2305)'} is assessed to ensure that the train's schedule matches the action's timing requirements. In the simulator, specific scoring items are designated to calculate the score. The scoring for Level 1 is calculated as follows:
\begin{equation}
s_1 = w_1 \times \frac{A_1}{B_1},
\end{equation}
where \( w_1 \) is the weight of Level 1 (60 points), \( A_1 \) is the number of successfully completed scoring items, and \( B_1 \) is the total number of scoring items in Level 1.

\textbf{Level 2 - Constraint Satisfaction.} This level assesses the itinerary's adherence to specified constraints like budget limits, stay durations, and opening hours of tourist spots. The scoring for Level 2 is calculated as follows:
\begin{equation}
s_2 = w_2 \times \frac{A_2}{B_2},
\end{equation}
where \( w_2 \) is the weight of Level 2 (20 points), \( A_2 \) is the number of successfully completed scoring items, and \( B_2 \) is the total number of scoring items in Level 2.

\textbf{Level 3 - Time and Cost Efficiency.} This level evaluates the itinerary based on its efficiency in terms of time and cost, while also considering the criteria from the previous levels. Efficiency is assessed against a predefined spending range \([a, b]\). A higher score is awarded for lower expenditures within this range, with a full score for expenses at or below the lower limit \(a\), and no score for exceeding the upper limit \(b\). The scoring formula is as follows:
\begin{equation} s_3 = \begin{cases} 
w_3 & \text{if } s \leq a \\
w_3 \times \left(1 - \frac{s - a}{b - a}\right) & \text{if } a < s < b\,\,\,\,\,, 
 \\
0 & \text{if } s \geq b
\end{cases} \end{equation}
where \( w_3 \) is the weight of Level 3 (20 points) and \( s \) is the actual cost or time spent.

The values of \(a\) and \(b\) are determined by generating a large number of candidate itineraries and executing them within the simulator. This ensures that only the correct and executable itineraries are considered. From this pool, we randomly select 50 valid itineraries, and their time and cost values are used to calculate the mean and standard deviation. We set \(a = \mu - \sigma\) and \(b = \mu + \sigma\), where \( \mu \) is the mean and \( \sigma \) is the standard deviation. This ensures that the range \([a, b]\) reflects realistic and achievable time and cost efficiency based on the performance of valid itineraries in the simulator.

These metrics provide a nuanced assessment of an agent's performance, especially in scenarios where only partial task completion is achieved. This approach ensures a more accurate reflection of an agent's capabilities in handling complex and layered tasks.

\noindent\textbf{Sequential Evaluation Process.} It is important to note that the evaluation for higher levels only proceeds after the complete attainment of scores in the lower levels. For instance, an itinerary must first be feasible (Level 1) before it can be evaluated for compliance with specific constraints (Level 2) and its efficiency in terms of time and cost (Level 3) This approach reinforces the importance of a strong base in practical execution, as advanced optimizations are irrelevant if basic executability is not achieved.

\subsection{Feature Highlights}
\label{sec:features}
The features of \dataset are summarized as follows:

\begin{enumerate}
  \item \textbf{Multi-Faceted and Fine-Grained Evaluation.} Moving beyond binary success-failure assessments, our benchmark employs a nuanced scoring system, which recognizes varying degrees of task completion and evaluates agents at multiple levels, reflecting their overall performance in realistic applications.
  
  \item \textbf{Interconnected Tasks and Progressive Complexity.} The tasks center around travel planning and are designed to increase in complexity through a series of evolving objectives and constraints. Each task is an extension of the previous one, simulating the interconnected nature of real-world problem-solving.
  
  \item \textbf{Integration of Tools.} The benchmark incorporates tools like the Python interpreter and the database to evaluate agents' abilities to utilize tools and handle external information. The Python interpreter can be used for planning optimization and the database provides access to additional information, such as transportation and attraction details. 
\end{enumerate}

\begin{figure*}[htbp]
    \centering
    \includegraphics[width=\textwidth]{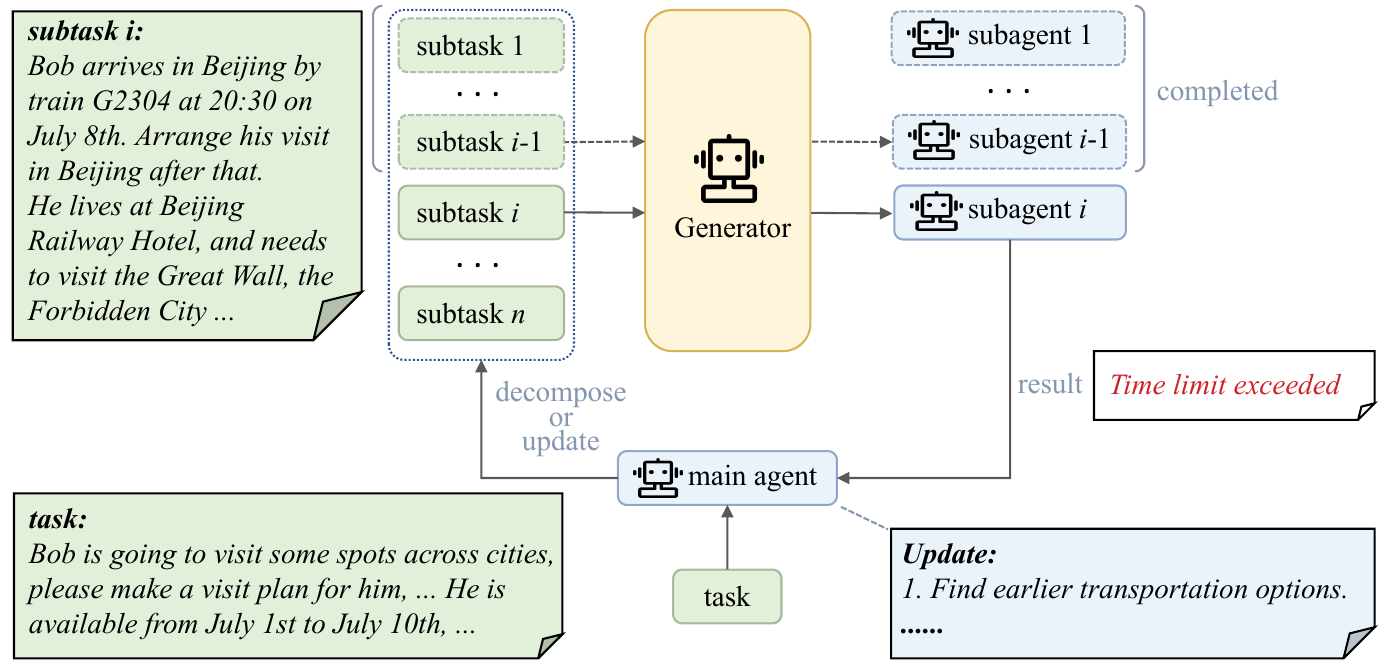}
    \caption{\textbf{Overview of TDAG framework.} It shows the process of decomposing a complex task into multiple subtasks, which are then dynamically updated based on the completion status of preceding subtasks. Each subtask is assigned to a specially generated subagent, ensuring targeted and efficient task execution. }
    \label{fig:framework-overview}
\end{figure*}

\section{A Multi-agent Framework based on Dynamic Task Decomposition and Agent Generation}
In this section, we introduce a multi-agent framework based on dynamic Task Decomposition and Agent Generation (TDAG), as illustrated in Figure~\ref{fig:framework-overview}. The framework is structured around two key components: (1) an advanced task decomposition strategy that dynamically adapts to the evolving requirements of complex tasks (Section~\ref{sec:task-decomposition}), and (2) a flexible agent generation process that customizes agents for specific subtasks, enhancing adaptability and efficiency in varied scenarios (Section~\ref{sec:agent-generation}).

\subsection{Dynamic Task Decomposition}
\label{sec:task-decomposition}
In our framework, complex tasks are decomposed into smaller, more manageable subtasks, addressing the challenge of excessive irrelevant contexts that may hinder LLM performance.

The main agent decomposes the task, and each subagent is assigned a subtask, the process of which is represented as:
\begin{equation}(t_1,t_2,\ldots,t_n)=\text{MainAgent}(T),\end{equation}
where \(t_i\) is the \(i\)-th subtask and \(T\) is the original task. 

The \(i\)-th subtask is performed by the corresponding \(\text{subagent}_i\), with the result as
\begin{equation}r_i=\text{SubAgent}_i (t_1,t_2,\ldots,t_i, r_1,r_2,\ldots,r_{i-1}).\end{equation}
Note that each subagent focuses on a specific subtask, reducing the potential for noise of irrelevant information, thereby enhancing performance.

Crucially, these decomposed subtasks are not static, but will be dynamically adjusted based on the outcomes of preceding tasks, which is represented as:
\begin{equation}t_i'=\text{Update}(t_i,r_1,r_2,\ldots,r_{i-1}),\end{equation}
where \(t_i'\) is the updated \(i\)-th subtask, redefined based on the results \(r_1,r_2,\ldots,r_{i-1}\) of preceding subtasks. This dynamic adjustment accounts for unexpected results (such as failures) or new information obtained during the execution of earlier subtasks, allowing for real-time recalibration and enhancement of the task-solving strategy.

\begin{algorithm*}[ht]
\caption{Multi-Agent Framework Operation}
\label{alg:framework-operation}
\begin{algorithmic}[1] 
\State \textbf{Input:} Task $T$, Tool document $D$, Skill library $L$
\State \textbf{Output:} Updated Skill library $L$, Task results $R$

\State $t\_list \leftarrow \text{MainAgent.Decompose}(T)$ \Comment{Decompose task T}
\For{$t_i$ in $t\_list$}
    \State $subagent_i \leftarrow \text{AgentGenerator.Generate}(D, L, t_i)$ \Comment{Generate subagent}
    \State $r_i \leftarrow subagent_i.Execute(t_1, t_2, \ldots, t_{i-1}, t_i, r_1, r_2, \ldots, r_{i-1})$ \Comment{Execute subtask}
    \If{$t_i$ is successfully completed}
        \State $s \leftarrow subagent_i.SummarizeProcess()$ \Comment{Summarize process}
        \State $L \leftarrow \text{Processor.UpdateSkillLibrary}(L, s)$ \Comment{Update Skill library}
    \EndIf
    \State $t\_list \leftarrow \text{MainAgent.UpdateTasks}(t\_list, r_i)$ \Comment{Update task list}
\EndFor
\State $R \leftarrow \text{MainAgent.Submit}(t_1, t_2, \ldots, t_n, r_1, r_2, \ldots, r_n)$ \Comment{Get the final result}
\State \Return $L, R$ \Comment{Return updated library and results}
\end{algorithmic}
\end{algorithm*}

\subsection{Agent Generation}
\label{sec:agent-generation}

Differing from previous methods \cite{planandexecute, adapt}, where tasks are assigned to predefined, fixed subagents, our framework allows for dynamic customization of subagents. This customization is achieved through LLM prompting, reflected in two aspects: generation of the tool document and an incremental skill library. By doing so, our framework significantly reduces the human effort required for agent setup and adaptation, making it more efficient in handling complex tasks.

\subsubsection{Tool Document Generation}

In standard tool documents, there are often issues of redundancy, unclear descriptions, which may mislead the agent into making inappropriate attempts to handle the task. Our framework involves generating an enhanced version of the original document by restructuring, clarifying and enriching the original content. By doing so, we aim to ensure that every piece of information presented to the agent is not only relevant but also optimally formatted for effective comprehension and application, thereby enhancing the agent's efficiency in task execution.

\subsubsection{Incremental Skill Library}

Our framework generates skills for subagents based on the agent behavior during task execution. These skills are then stored in a skill library, available for reference when agents face similar objectives in future tasks. The concept of a skill library is inspired by voyager \cite{voyager}, but with crucial adaptations. Unlike the environment of voyager, where the correctness of skills are directly verifiable through environmental feedback, our benchmark environment cannot guarantee the correctness of a skill. Therefore, the skills our agents develop are subject to continuous refinement in response to variable tasks and environmental conditions. When employing a skill, we use a small SentenceBERT model to retrieve the most relevant skill from the library based on the subtask's content. Once a subagent completes a subtask, it is prompted to summarize its process, which is then used to update the skill library.
The format for skills is as follows:

\begin{itemize}
    \item \textbf{Name:} The name of the skill.
    \item \textbf{Detail:} The details of the subtask.
    \item \textbf{Solution:} The solution for the subtask.
\end{itemize}

An agent dedicated to skill modification oversees the library, updating existing skills based on new data and experiences. This continuous cycle of generation, application, and refinement ensures that the skills in our library remain effective and adaptable to a wide range of scenarios. Algorithm~\ref{alg:framework-operation} illustrates the step-by-step process of the overall framework.

\section{Experiments}

\subsection{Settings}
Our experiments are conducted across the three types of tasks of the Itinerary Bench as mentioned in Section~\ref{sec:benchmark}. We employ the fine-grained evaluation system previously described, which allows scoring for partial completions of tasks. The baseline methodologies we have employed are:

\begin{itemize}
  \item \textbf{ReAct} \citep{react}. A single agent iteratively completes the task, with each step involving the agent giving out thoughts and actions based on current observations.

  \item \textbf{P\&S} \citep{planandsolve}. In this setting, the agent first creates a plan with multiple steps and then follows the ReAct pattern to solve the problem step by step.

  \item \textbf{P\&E} \citep{planandexecute}. The task is initially broken down into multiple subtasks, which are then sequentially assigned to a predefined executor.

  \item \textbf{ADAPT} \citep{adapt}. This method employs a recursive strategy to break down tasks into subtasks, which are then assigned to specific subagents. Subtasks can be further decomposed if not completed successfully.
\end{itemize}

To further explore the contributions of the dynamic decomposition and agent generation components in our TDAG framework, we conducted ablation studies by individually removing these features. Specifically, we evaluated the performance of our model without the dynamic decomposition process, using a static planner as in P\&E method, and without the customization of subagents, relying instead on predefined subagents.

\subsection{Main Results}

In our experiments, we use GPT-3.5-turbo-16k as the foundational model, and the results are shown in Table~\ref{tab:main-results}. Our method, TDAG, demonstrates superior performance compared to the established baselines. Furthermore, the ablation results 
confirm that the integration of both dynamic task decomposition and agent generation is crucial for optimizing task execution in complex, unpredictable environments.

Interestingly, the P\&E approach, while allocating specific subtasks to designated subagents, does not outperform ReAct, where a single agent executes the entire task. This observation can be primarily attributed to the fixed nature of task assignments in P\&E, as discussed in Section~\ref{subsec:error-analysis}. Once subtasks are assigned, subagents lack the capacity to alter the plan, leading to challenges when initial subtasks do not produce expected results. This rigidity can make subsequent subtasks unexecutable, which is a critical limitation in dynamic and unpredictable environments. On the other hand, TDAG maintains the flexibility to adjust its execution strategy based on evolving task conditions, a key advantage in complex environments. 

\begin{table*}[t]
    \centering
    \begin{tabular}{l|cccc}
        \toprule
        \textbf{Method} & \textbf{Type 1} & \textbf{Type 2} & \textbf{Type 3} & \textbf{Avg.} \\ 
        \midrule
        ReAct & 43.84 & 42.69 & 42.54 & 43.02\\
        P\&S & 41.28 & 46.48 & 43.27 & 43.68  \\
        P\&E & 39.09 & 47.44 & 42.03 & 42.85 \\
        ADAPT &42.73 & 48.58 & 42.92 & 44.74  \\
        \midrule
        TDAG (Ours) &\textbf{49.78} & \textbf{50.96} & \textbf{46.51} & \textbf{49.08 } \\
        \quad w/o Agent Generation & 47.2 & 47.1 & 45.78 & 46.69 \\
        \quad w/o Dynamic Decomposition & 44.7 & 50.04 & 43.94 & 46.23 \\
        \bottomrule
    \end{tabular}
    \caption{Experimental results for each method on three task types of ItineraryBench. Numbers in bold indicate the highest score among all methods.}
    \label{tab:main-results}
\end{table*}

\begin{table}[t]
    \centering
    \begin{tabular}{lcccc}
        \toprule
        \textbf{Method} & \textbf{CTF} & \textbf{CKE} & \textbf{EIM} & \textbf{CNC} \\
        \midrule
        ReAct           & 32.61 & 19.75 & 17.32 & 20.34 \\
        P\&S            & 8.70 & 21.97 & 19.46 & 21.91 \\
        P\&E            & 34.78 & 20.94 & 22.14 & 21.05 \\
        ADAPT           & 19.57 & 18.47 & 21.74 & 18.07 \\
        \midrule
        TDAG            & 4.35 & 18.87 & 19.33 & 18.63 \\
        \bottomrule
    \end{tabular}
    \caption{Percentage of each type of errors encountered by different methods. The values in the table represent the percentage of each type of errors made by a method relative to the total number of that type of errors.}
    \label{tab:error_analysis}
\end{table}

\begin{figure}[ht]
    \centering
    \includegraphics[width=\linewidth]{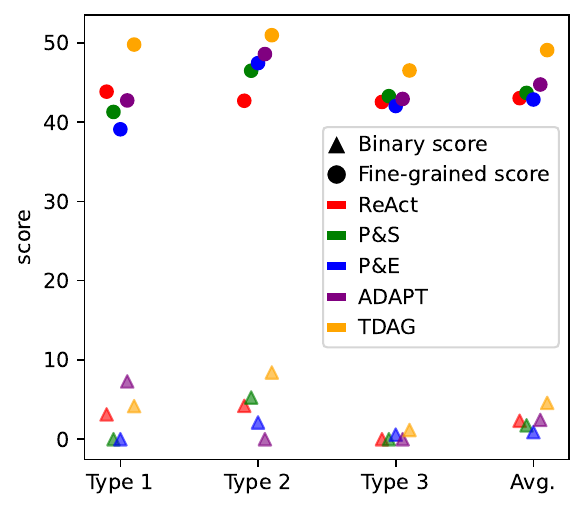}
    \caption{Comparison of method performance using binary scoring and fine-grained evaluation.}
    \label{fig:evaluation-figure}
\end{figure}

\begin{table}[htbp]
\centering
\resizebox{\linewidth}{!}{%
\begin{tabular}{cccccc}
\toprule
\multirow{2}{*}{\textbf{Method}} & \multicolumn{2}{c}{\textbf{WebShop}} & \multicolumn{1}{c}{\textbf{TextCraft}} \\
\cmidrule(lr){2-3} \cmidrule(lr){4-4}
 & reward score & success rate & success rate \\
\midrule
ReAct & 42.1 & 29.0 & 19.0 \\
P\&S & 58.2 & 23.0 & 18.5 \\
P\&E & 27.7 & 17.0 & 27.0 \\
ADAPT & 60.0 & 44.0 & 52.0 \\
\midrule
TDAG & \textbf{64.5} & \textbf{45.0} & \textbf{73.5} \\
\bottomrule
\end{tabular}%
}
\caption{Experimental Results for WebShop and TextCraft. Following ~\citet{adapt}, we report the reward score and success rate (\%) for WebShop and success rate for TextCraft.}
\label{tab:combined_results}
\end{table}

\subsection{Error Analysis}
\label{subsec:error-analysis}
In our analysis, we identified and categorized the primary error types encountered by the different methods. These error types are as follows:

\begin{itemize}
    \item \textbf{Cascading Task Failure (CTF)}: The entire task fails due to the failure of intermediate subtasks, often as a result of error propagation.
    \item \textbf{Commonsense Knowledge Error (CKE)}: Errors in basic understanding of real-world common sense, such as mismatch between the origin in \texttt{go\_to\_place} and the character's current location.
    \item \textbf{External Information Misalignment (EIM)}: Errors in using external information, mainly stemming from the hallucination phenomenon~\citep{hallucination}, such as inconsistencies between database-provided ticket information and the travel plan.
    \item \textbf{Constraint Non-compliance (CNC)}: Errors related to not adhering to specific constraints of the task. This includes exceeding budget or time limits, etc.
\end{itemize} 

As represented in Table~\ref{tab:error_analysis}, methods with fixed plans, such as P\&E, are particularly prone to Cascading Task Failures, suggesting a lack of flexibility in adapting to unforeseen issues within a task.  This type of error, where an early mistake can lead to the failure of an entire sequence of tasks, underscores the critical need for dynamic task decomposition. Regarding other types of errors, our analysis indicates that the TDAG method generally exhibits low error rates across CKE, EIM and CNC errors. This improvement is attributed to the ability of dynamically generated subagents to focus on specific subtasks, leveraging their tailored capabilities to navigate the intricacies of each challenge effectively. Notably, ReAct outperforms more complex approaches in minimizing EIM errors. A possible reason could be that complex methods generate excessive context, which may increase the likelihood of hallucination phenomena.

\subsection{Binary Scoring v.s. Fine-grained Evaluation}
To highlight the significance of the fine-grained evaluation, we conducted an experiment comparing it with the traditional binary scoring method. Given the complexity of tasks, a full score in Level 1 - Executability was used as the criterion for successful task completion, with the results presented in Figure~\ref{fig:evaluation-figure}. It reveals that in challenging tasks with a low probability of completion, binary scoring fails to distinguish between different methods effectively. On the other hand, the fine-grained evaluation method provides a more detailed insight, reflecting the extent of task completion even when perfect execution is not achieved.

\subsection{Generalizability}

To further assess the generalizability of our TDAG method, we conducted additional experiments on the following datasets:
\begin{itemize}
    \item \textbf{WebShop}~\cite{webshop}. Developed as a simulated e-commerce environment, WebShop requires agents to navigate through various webpage types, reformulate queries, comprehend and interact with text-heavy webpages. Following ~\citet{reflexion}, we report performance on 100 user instructions.
    \item \textbf{TextCraft}~\cite{adapt}. As a text-only environment mirroring the crafting component of Minecraft, TextCraft presents tasks with a natural compositional structure akin
    . Agents in TextCraft aim to craft target items using available resources and a set of defined actions. The test split contains 200 samples.
\end{itemize}
Following ~\citet{adapt}, we employed GPT-3.5-turbo for WebShop and gpt-3.5-turbo-instruct for TextCraft. Due to the lack of tool document and the monotonous nature of the tasks, we did not engage in agent generation. Instead, we utilized the predefined agents from ADAPT.

The experimental results, as illustrated in Table~\ref{tab:combined_results}, demonstrate that our method outperforms the baseline methods in both reward score and success rate metrics. The results reinforce our assertion that TDAG is not only effective in complex, multifaceted tasks but also excels in more monotonous and straightforward scenarios.
\section{Conclusion}
In this paper, we first present \dataset, a benchmark for evaluating LLM-based agents in complex, real-world tasks, particularly in travel planning. It stands out with its interconnected, progressively challenging tasks and a nuanced evaluation system that goes beyond binary scoring. This approach allows for a more precise assessment of an agent's capabilities, especially in partial task completions. Additionally, we introduce the TDAG framework, enhancing adaptability and success in diverse tasks by dynamically decomposing complex tasks into manageable subtasks, each handled by a custom-generated subagent. Our experimental results show that TDAG significantly outperforms existing baselines across several benchmarks.
\section*{Acknowledgment}
The project was supported by 
National Natural Science Foundation of China (No. 62036004, No. 62276219),  
Natural Science Foundation of Fujian Province of China (No. 2024J011001),
and
the Public Technology Service Platform Project of Xiamen (No.3502Z20231043).
We also thank the reviewers for their insightful comments.
\bibliography{custom}

\appendix
\section{Benchmark Details}
\label{sec:appendix}

In this appendix, we provide detailed insights into the composition and characteristics of the \dataset, focusing on the distribution of task types, city and attraction number distributions, and the various constraints that define the travel planning scenarios.

\subsection{Task Type Distribution}

Table~\ref{tab:type_distribution} presents the distribution of task types within the benchmark. The tasks are categorized into three types, each representing a distinct travel planning scenario. Type 1 tasks involve inter-city travel planning, focusing on arranging city visits and transportation. Type 2 tasks are centered around intra-city navigation, requiring planning visits to various attractions within a single city. Type 3 combines elements of both Type 1 and Type 2, presenting a more complex scenario of navigating through multiple cities and their attractions. The distribution of these task types is designed to ensure a comprehensive evaluation of agents' capabilities across different planning complexities.

\begin{table}[h]
    \centering
    \begin{tabular}{lc}
        \toprule
        \textbf{Type} & \textbf{Number} \\
        \midrule
        Type 1 & 96 \\
        Type 2 & 95 \\
        Type 3 & 173 \\
        \bottomrule
    \end{tabular}
    \caption{Distribution of task types in \dataset.}
    \label{tab:type_distribution}
\end{table}

\subsection{City and Attraction Distribution}

\begin{figure}[ht]
\centering
\includegraphics[width=\linewidth]{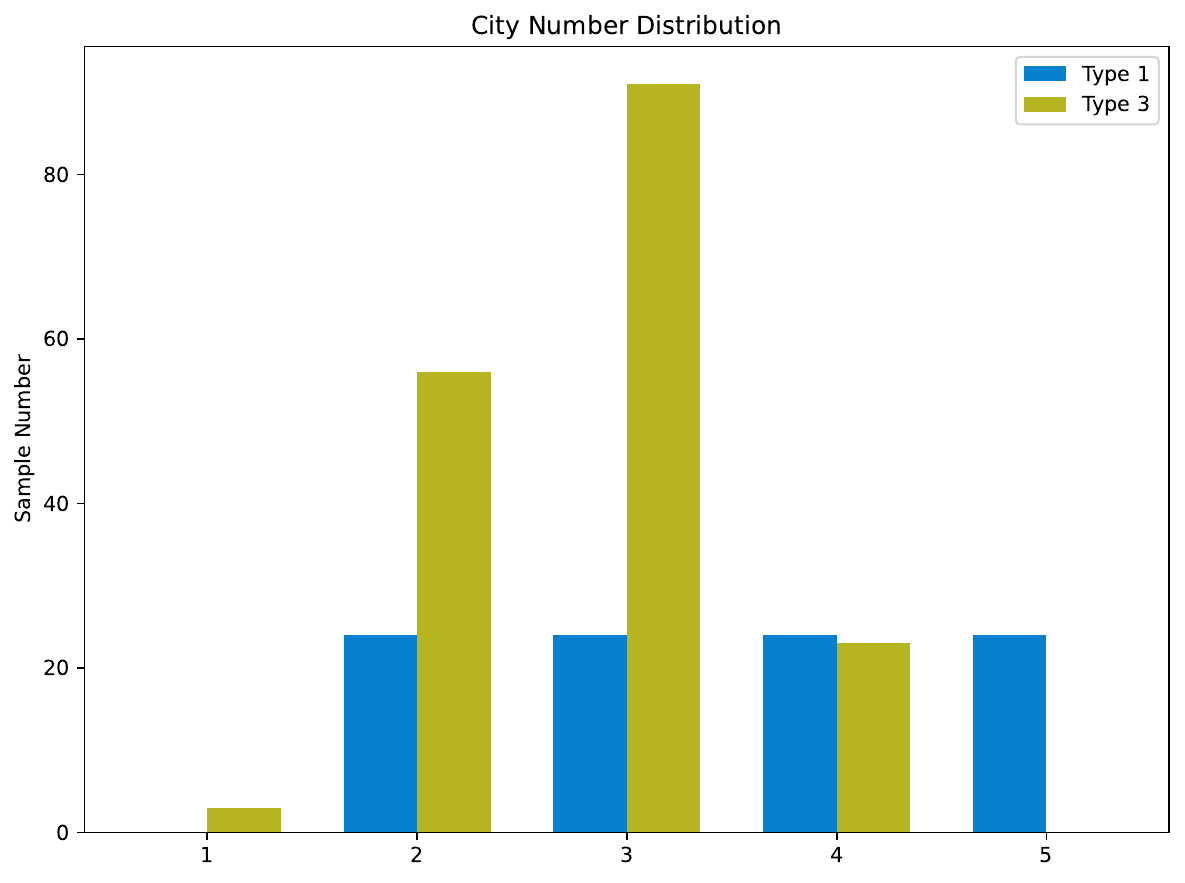}
\caption{City Number Distribution.}
\label{fig:city_distribution}
\end{figure}

\begin{figure}[ht]
\centering
\includegraphics[width=\linewidth]{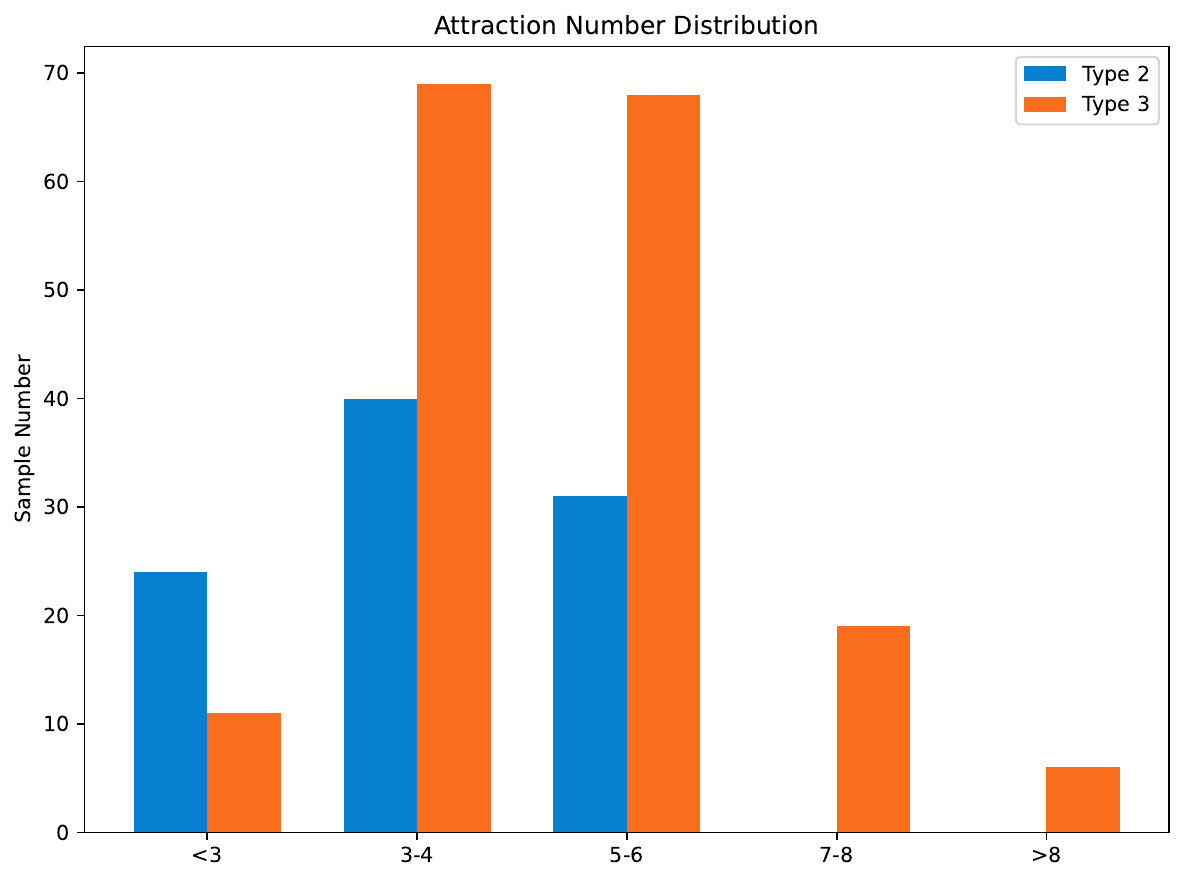}
\caption{Attraction Number Distribution.}
\label{fig:attraction_distribution}
\end{figure}

The distributions of cities and attractions within the dataset are visualized in Figures~\ref{fig:city_distribution} and \ref{fig:attraction_distribution}, respectively. These distributions highlight the benchmark's diversity and the range of scenarios that agents must navigate. The city distribution figure shows the number of times each city appears across the dataset, reflecting the geographical spread and frequency of visits required in various tasks. Similarly, the attraction distribution figure illustrates the variety and frequency of tourist spots that must be considered in the travel plans.

\subsection{Constraints in Travel Planning}

Table~\ref{tab:constraints} outlines the various constraints incorporated into the travel planning tasks. These constraints simulate real-world conditions and challenges, requiring agents to optimize their plans while adhering to limitations such as time, budget, transportation options, and specific requirements for city and spot durations. Including constraints such as specific hotel stays, designated activity times, and rest periods further enhances the benchmark's realism, pushing the boundaries of agents' problem-solving abilities. 

It is noteworthy that all attraction and transportation information were either obtained from public websites or self-edited, ensuring that there are no issues related to privacy, copyright, or ethics.

\begin{table*}[ht]
\centering
\begin{tabular}{lp{8cm}}
\toprule
\textbf{Constraint} & \textbf{Description} \\
\midrule
Time Limit & The travel must be completed within a certain period. \\
Budget & A budget for the trip. \\
Transportation & Includes both intra-city and inter-city transportation. \\
City Duration & A specific duration of stay in a city. \\
Spot Duration & A specific duration of stay at a tourist spot. \\
Specific Hotel & Staying in a specified hotel. \\
Activity Time & Designated times when the user is allowed to engage in activities. \\
Spot Opening Hours & The operating hours of tourist attractions. \\
Rest Time & Designated times when the user must rest. \\
\bottomrule
\end{tabular}
\caption{Types of Constraints in Travel Planning.}
\label{tab:constraints}
\end{table*}

\section{Experimental Details}
\subsection{Implementation of the Incremental Skill Library}

The Incremental Skill Library includes skill generation, storage, and retrieval. Skills are organized into three fields: `Name', `Detail', and `Solution'.
After completing a subtask, subagents use LLM prompting to summarize their solution. This process involves determining the uniqueness of the proposed skill to prevent redundancy in the library. The SentenceBERT model `all-mpnet-base-v2' is utilized to generate embeddings for the skill details. A similarity threshold, \(\theta = 0.7\), is set to identify potential duplicates. A new skill is considered redundant and excluded from the library if there are already \(k\) or more skills in the library with a similarity greater than \(\theta\), where  \(k=2\).
For retrieving skills, the same SentenceBERT model is employed. When a subagent is assigned a new subtask, the model finds the top \(k\) most relevant skills by comparing the embeddings of the subtask's name to those in the library. These skills are then provided to the subagent in textual form, aiding in the execution of the task.

\subsection{Models and Hyperparameters}

For the experiments on \dataset, we deploy the gpt-3.5-turbo-16k. Given the complex nature and the lengthy process required to solve tasks in this benchmark, methods lacking a task decomposition strategy, such as ReAct, require long context length to achieve task completion. For WebShop and TextCraft, we adhere to the configurations recommended by \citet{adapt}, utilizing the gpt-3.5-turbo and gpt-3.5-turbo-instruct respectively.
To ensure experimental consistency and replicability, we set the temperature to 0.

\end{document}